\documentclass[journal]{IEEEtran}
\IEEEoverridecommandlockouts
\usepackage{cite}
\usepackage{amsmath,amssymb}
\usepackage{graphicx}
\usepackage[dvipsnames]{xcolor}
\usepackage{url}
\usepackage{booktabs,colortbl}
\definecolor{tabfirst}{HTML}{FFB994}
\definecolor{tabsecond}{HTML}{FFE4D6}
\expandafter\def\expandafter\normalsize\expandafter{\normalsize\setlength\abovedisplayskip{5.5pt plus 1.5pt minus 2pt}\setlength\belowdisplayskip{5.5pt plus 1.5pt minus 2pt}}
\definecolor{cvprblue}{rgb}{0.21,0.49,0.74}
\usepackage[breaklinks,colorlinks,allcolors=cvprblue]{hyperref}
\makeatletter
\let\cosphere@maketitle\@maketitle
\renewcommand{\@maketitle}{\cosphere@maketitle
\vspace{12pt}
\begin{center}\includegraphics[width=\textwidth]{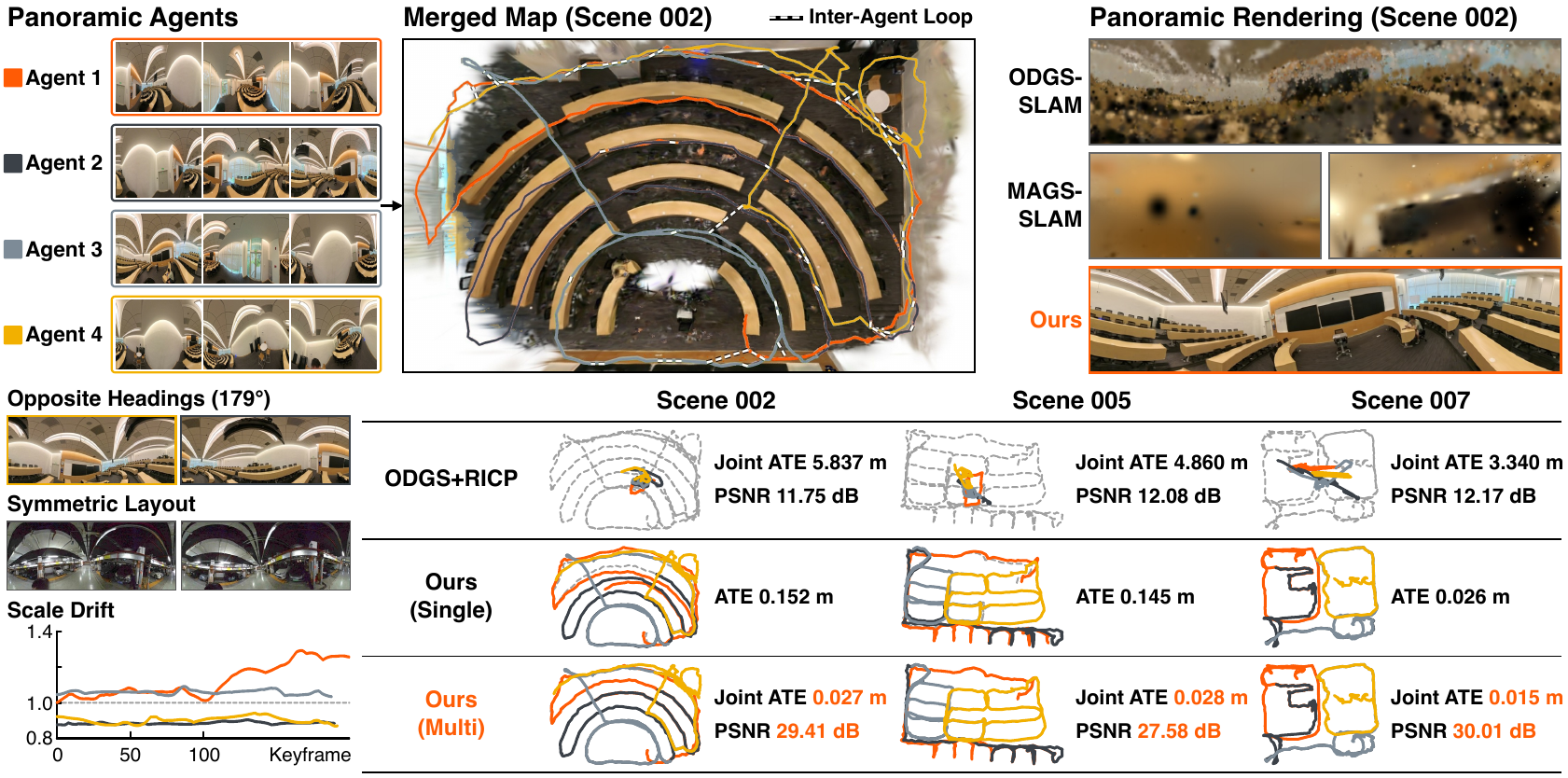}\end{center}
\vspace{-6pt}{\footnotesize\setcounter{figure}{0}\refstepcounter{figure}\label{fig:teaser}Fig.~\thefigure.\enskip Collaborative panoramic Gaussian splatting SLAM with CoSphere-SLAM. Four monocular panoramic agents at top left are aligned into one 3D Gaussian map at top center with $360^\circ$ rendering at top right on Scene 002. While the bottom left shows a loop closure at opposite headings, a symmetric layout, and scale drift, the bottom right compares trajectories, single-agent ATE or multi-agent joint ATE, and masked front-face PSNR across three development scenes against dashed ground truth, with RICP registering ODGS-SLAM maps via RANSAC and ICP.\par}
\vspace{8pt}}
\makeatother

\begin{document}
\title{CoSphere-SLAM: Multi-Agent Panoramic Gaussian Splatting SLAM with Spherical Loop Closure\vspace{8pt}}
\author{Qi Shao, Zhihao Cao, Shuhao Zhai, Taoyu Wu, Zhuoxiao Li, Anh Nguyen, Long Hu, Xianzhi Li, Hong Liang, Liang Li, and Baoru Huang%
\thanks{Corresponding authors: Liang Li and Baoru Huang.}%
\thanks{Q. Shao, H. Liang, and L. Li are with Harbin Engineering University, Harbin, China. Q. Shao is also with the University of Liverpool, Liverpool, UK.}%
\thanks{Z. Cao is with ETH Zurich, Zurich, Switzerland, and also with the National University of Singapore, Singapore.}%
\thanks{S. Zhai is with the University of Macau, Macau, China.}%
\thanks{T. Wu is with the University of Liverpool, Liverpool, UK, and also with Xi'an Jiaotong-Liverpool University, Suzhou, China.}%
\thanks{Z. Li is with The Hong Kong University of Science and Technology (Guangzhou), Guangzhou, China.}%
\thanks{A. Nguyen and B. Huang are with the University of Liverpool, Liverpool, UK, and also with Imperial College London, London, UK.}%
\thanks{L. Hu and X. Li are with Huazhong University of Science and Technology, Wuhan, China.}}
\markboth{}{}

\maketitle
\begin{abstract}
Dense collaborative mapping is essential for search and rescue, building inspection, and warehouse automation, where panoramic cameras capture the entire surroundings in a single frame. Unlike pinhole agents that share views only at similar headings, panoramic agents observe overlapping scenes at any 3D orientation, yet searching full rotations also produces mutually consistent false loop closures across distinct symmetric rooms. Moreover, continuous monocular scale drift splits rigidly merged maps at revisited surfaces, while depth noise at short baselines distorts loop translation directions. To address these challenges, we propose CoSphere-SLAM, the first multi-agent Gaussian splatting SLAM framework for panoramic cameras, whose spherical loop closure aligns spherical features over $\mathrm{SO}(3)$ via fast Fourier correlation across arbitrary orientations and rejects symmetric false closures through global $\mathrm{Sim}(3)$ consensus. Building on these closures, directional global optimization solves a per-keyframe $\mathrm{Sim}(3)$ pose graph with log-scale regularization to correct scale drift while separating optical-flow translation directions from depth-based magnitudes. Finally, omnidirectional Gaussian fusion transforms each Gaussian with its keyframe similarity and renders Gaussians at their maximum ray response to jointly refine poses and maps. Experiments on indoor and cross-session multi-agent panoramic benchmarks demonstrate that CoSphere-SLAM achieves state-of-the-art trajectory accuracy and rendering quality among single-agent and multi-agent methods. Project website: \href{https://cosphere-slam.github.io}{cosphere-slam.github.io}.
\end{abstract}
\section{Introduction}
Dense 3D mapping of large indoor environments is essential for search and rescue, building inspection, and warehouse automation, where 3D Gaussian splatting \cite{kerbl20233dgs} enables dense photorealistic SLAM from a moving camera \cite{keetha2024splatam,matsuki2024monogs,huang2024photoslam}. Equipping an agent with a $360^\circ$ panoramic camera is particularly effective indoors, since each frame captures the whole room in one pass and constrains motion in all directions \cite{lee2024odgs,li2025omnigs,mao2026panogsslam,guo2026cubesplat,spiss2026odgsslam}. Even with $360^\circ$ views, however, a single agent observes the scene along only one path, where walls and furniture occlude surfaces behind them, visiting rooms one by one is slow, and monocular scale and pose drift accumulate across rooms. Mapping large indoor spaces rapidly and completely therefore requires extending single-agent panoramic SLAM to multiple agents exploring parallel routes and sessions \cite{schmuck2019ccm,lajoie2020door,tian2022kimera}. Yet existing multi-agent systems \cite{hu2023cpslam,deng2024mne,yugay2025magic,xu2025macego3d,mags2025,como3r} rely on narrow pinhole views and RGB-D inputs or rigid submaps, missing meetings across different headings on cropped views while failing under equirectangular distortion, symmetric rooms, and scale drift on panoramas.

Extending panoramic Gaussian SLAM to multiple agents must resolve three coupled challenges across loop closure, global optimization, and map fusion. First, as mobile cameras pitch and roll across arbitrary headings, yaw shifts an equirectangular image cyclically along longitude, whereas tilt warps rows differently across latitudes and breaks planar matching. Moreover, searching full 3D rotations matches symmetric rooms up to a rigid transform, and unlike random outliers, all matches in a symmetric layout imply the same wrong transform between agents, so pairwise robust kernels cannot reject them. Second, monocular trajectories drift continuously in scale between distant meetings, whereas depth errors at short baselines shift 3D point alignment in proportion to scene depth and corrupt translation directions. Third, varying scale corrections split rigidly merged submaps at revisited walls, while projecting 3D Gaussians onto a tangent plane stretches them near the poles and cuts across the seam.

A monocular panorama observes the unit sphere $\mathcal{S}^2$ without field-of-view limits, providing scale-free viewing directions while leaving discrete room symmetries and continuous radial scale unobservable. Whereas $\mathrm{SO}(3)$ search turns symmetric rooms into coherent outlier clusters and isotropic alignment conflates flow directions with depth noise, \textbf{CoSphere-SLAM} couples three $\mathcal{S}^2$ designs to resolve both ambiguities without distortion (Figure~\ref{fig:teaser}). Building on per-agent full-sphere tracking and ray-based mapping, \textit{spherical loop closure} correlates features over $\mathrm{SO}(3)$ at arbitrary orientations and enforces cycle-consistent global $\mathrm{Sim}(3)$ consensus across agent loops to reject symmetric clusters. \textit{Directional global optimization} then solves an anisotropic per-keyframe $\mathrm{Sim}(3)$ pose graph separating flow directions from scale drift, and \textit{omnidirectional Gaussian fusion} deforms Gaussians by keyframe similarities to refine poses and maps along viewing rays.

Our main contributions are as follows:
\begin{itemize}\setlength{\itemsep}{1pt}\setlength{\parskip}{0pt}\setlength{\topsep}{2pt}
\item We design the first multi-agent panoramic Gaussian splatting SLAM framework, CoSphere-SLAM, unifying full-sphere camera tracking, compact descriptor communication, and dense mapping without prior poses or scales.
\item We propose spherical loop closure to correlate features over $\mathrm{SO}(3)$ at arbitrary orientations and reject symmetric false closures via global $\mathrm{Sim}(3)$ consensus.
\item We introduce directional global optimization to separate optical-flow translation directions from monocular scale drift on a per-keyframe $\mathrm{Sim}(3)$ pose graph.
\item We develop omnidirectional Gaussian fusion to align local Gaussians with per-keyframe similarities and jointly refine poses and maps along viewing rays.
\end{itemize}
\section{Related Work}
\label{sec:related}
\noindent\textbf{Multi-Agent SLAM.} Collaborative SLAM merges agent trajectories through inter-agent loop closures. Centralized systems \cite{schmuck2019ccm,schmuck2021covins} optimize a global map on a server, whereas distributed systems \cite{lajoie2020door,tian2022kimera,lajoie2024swarm} exchange compact descriptors and solve the pose graph across agents. To reject false inter-agent closures, these systems use pairwise consistency maximization \cite{mangelson2018pcm}, graduated non-convexity \cite{yang2020gnc}, and robust kernels such as dynamic covariance scaling \cite{agarwal2013dcs}, while cycle consistency over pose loops resolves symmetric structures in structure from motion \cite{zach2010disambiguating}. However, pairwise kernels reject only mutually inconsistent outliers, and pinhole views connect agents only at similar headings.

\noindent\textbf{Dense Collaborative Mapping.} Beyond single-agent Gaussian splatting SLAM, feed-forward models now reconstruct long monocular sequences \cite{deng2025vggtlong,chen2026ttt3r}. For multiple agents, CP-SLAM \cite{hu2023cpslam} shares neural point clouds between two agents, MNE-SLAM \cite{deng2024mne} distributes a neural implicit map, and MAGiC-SLAM \cite{yugay2025magic} and MAC-Ego3D \cite{xu2025macego3d} merge Gaussian splatting maps of RGB-D agents. MAGS-SLAM \cite{mags2025} extends Gaussian splatting to monocular agents with submap-level $\mathrm{Sim}(3)$ alignment, and CoMo3R-SLAM \cite{como3r} builds dense multi-agent maps with 3D reconstruction priors. All of them, however, rely on pinhole images and connect agents only where narrow views overlap, whereas panoramas share content whenever paths pass nearby.

\noindent\textbf{Panoramic SLAM and Reconstruction.} Panoramic cameras observe the full sphere and therefore constrain camera motion in all directions. OpenVSLAM \cite{sumikura2019openvslam} and 360VO \cite{huang2022360vo} track equirectangular images with sparse features and direct alignment, while omnidirectional Gaussian splatting reconstructs scenes from posed or unposed panoramas \cite{lee2024odgs,li2025omnigs,zhuang2026pfgs360} and extends to single-agent panoramic video SLAM \cite{mao2026panogsslam,guo2026cubesplat,spiss2026odgsslam}. In multi-robot estimation, Omni-swarm \cite{xu2022omniswarm} and D$^2$SLAM \cite{xu2024d2slam} use omnidirectional cameras in aerial swarms, yet estimate only sparse states without loop closure, scale correction, or dense mapping.
\section{Method}
\label{sec:method}
\begin{figure*}[t]
\centering
\includegraphics[width=1\linewidth]{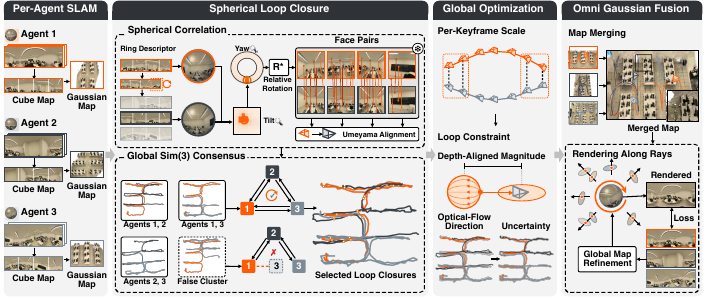}
\caption{Overview of CoSphere-SLAM. Each agent resamples panoramas into six cube-map faces with scale-aligned depth to track poses and build local 3D Gaussians. On the server, \textit{spherical loop closure} aligns spherical feature maps over $\mathrm{SO}(3)$ via circular FFT and removes symmetric false closures via global $\mathrm{Sim}(3)$ consensus across clusters and agent loops. \textit{Directional global optimization} solves a per-keyframe $\mathrm{Sim}(3)$ pose graph separating optical-flow directions from depth-aligned magnitudes, and \textit{omnidirectional Gaussian fusion} deforms Gaussians by their keyframe similarities to refine all poses and maps along viewing rays.}
\label{fig:pipeline}
\end{figure*}
As shown in Figure~\ref{fig:pipeline}, given monocular equirectangular streams from multiple panoramic agents with unknown initial poses and scales, CoSphere-SLAM estimates a camera-to-world similarity $\mathbf{S}_i=(s_i,\mathbf{R}_i,\mathbf{t}_i)\in\mathrm{Sim}(3)$ per keyframe $i\in\mathcal{V}$ and one unified 3D Gaussian map $\mathcal{G}$. While each panorama observes the full sphere $\mathcal{S}^2$ of viewing directions, monocular scale and room symmetry remain unobservable within one view. Building on local full-sphere tracking and mapping per agent (Sec.~\ref{sec:agent}), a server therefore aligns spherical features over $\mathrm{SO}(3)$ under global $\mathrm{Sim}(3)$ consensus (Sec.~\ref{sec:assoc}), optimizes per-keyframe scales while decoupling loop directions from scale drift (Sec.~\ref{sec:pgo}), and fuses all maps along viewing rays (Sec.~\ref{sec:gsmap}).

\subsection{Per-Agent Tracking and Mapping}
\label{sec:agent}
An equirectangular image stretches pixels near the poles and cuts across a vertical seam, preventing standard 2D convolutions from treating all directions equally. Each agent therefore resamples every panorama into six $90^\circ$ cube faces sharing one optical center and fixed body-to-face rotations $\mathbf{R}_f\in\mathrm{SO}(3)$, tracking all faces jointly with a DROID-SLAM front end \cite{teed2021droid} via $\mathbf{T}_i\mathbf{R}_f^{\top}$ to estimate one camera-to-local pose $\mathbf{T}_i\in\mathrm{SE}(3)$ per keyframe $i$. Predicting monocular depth on wider overlapping faces aligns per-face log-scales via linear least squares into one spherical depth map $D_i:\mathcal{S}^2\rightarrow\mathbb{R}_+$. Since predicted depth scale shifts across rooms, however, $D_i$ constrains pose updates only on the first keyframes while later keyframes rely solely on multiview optical flow and intra-agent closures. Holding $\mathbf{T}_i$ fixed then refines keyframe depths against $D_i$ to supervise local 3D Gaussians $\mathcal{G}_i$ along viewing rays (Sec.~\ref{sec:gsmap}). Meanwhile, averaging horizontal face features across rows forms a compact ring descriptor $\mathbf{H}_i\in\mathbb{R}^{C\times W_r}$ sent with $\mathbf{T}_i$.

\subsection{Spherical Loop Closure}
\label{sec:assoc}
On the server, inter-agent loop closure must match keyframes even when two agents pass at different headings and tilts. Resampling the dense features of the six cube faces onto an $H\times W$ equirectangular grid of unit directions defines a spherical feature map $F_i:\mathcal{S}^2\rightarrow\mathbb{R}^C$, under which nearby keyframes $i$ and $j$ observe the same sphere up to a coarse rotation $\mathbf{R}_{ij}\in\mathrm{SO}(3)$:
\begin{equation}
F_j(\mathbf{x})\approx F_i(\mathbf{R}_{ij}\,\mathbf{x})\quad\text{for all}\ \mathbf{x}\in\mathcal{S}^2.
\label{eq:sphere}
\end{equation}
Whereas rotating a pinhole camera crops the scene, rotating a panorama permutes $\mathcal{S}^2$ without loss and initializes coarse $\mathbf{R}_{ij}$ under small-baseline parallax.

To screen all $|\mathcal{V}|^2$ pairs, pooling horizontal face features over image rows makes $\mathbf{H}_i$ tolerant to tilt while turning yaw into a cyclic column shift $\Pi_v$. The server scores pairs by circular cross-correlation $\rho_{ij}=\max_v\langle\mathbf{H}_i,\Pi_v\mathbf{H}_j\rangle$ and requests faces and depths only for top pairs.

\noindent\textbf{Spherical Correlation.} When a handheld or mobile camera pitches or rolls, however, scene points move across latitudes of the equirectangular grid and 1D column shifts no longer align $F_i$ and $F_j$. To avoid a brute-force search over three Euler angles on $\mathrm{SO}(3)$, we factorize $\mathbf{R}=\mathbf{Q}\,\mathbf{R}_z(\psi)$ into a 2D tilt $\mathbf{Q}$ about a horizontal axis followed by a yaw $\mathbf{R}_z(\psi)$ about the vertical axis. Under a fixed tilt $\mathbf{Q}$, resampling $F_i$ into $F_i^{\mathbf{Q}}(\mathbf{x})=F_i(\mathbf{Q}\mathbf{x})$ turns correlation over all yaws $\psi\in[0,2\pi)$ into a 1D circular cross-correlation along longitude $\varphi$, evaluated by one inverse fast Fourier transform $\mathcal{F}_{\varphi}^{-1}$:
\begin{equation}
C_{ij}(\mathbf{Q},\cdot)=\mathcal{F}_{\varphi}^{-1}\!\Big[\sum_{\theta}\cos\theta\,\hat{F}^{\mathbf{Q}}_{i}(\theta,\cdot)^{\top}\overline{\hat{F}_{j}(\theta,\cdot)}\Big],
\label{eq:corr}
\end{equation}
where $\hat{F}=\mathcal{F}_{\varphi}[F]$ is the 1D discrete Fourier transform along $\varphi$ and $\cos\theta$ weights each cell by its solid angle at latitude $\theta$. Coarse-to-fine search over $N_Q$ tilts with parabolic yaw refinement initializes $\mathbf{R}^{*}_{ij}=\mathbf{Q}^{*}\mathbf{R}_z(\psi^{*})$ across $360^\circ$ headings in $\mathcal{O}(N_Q C H W\log W)$ time.

Initializing from $\mathbf{R}^{*}_{ij}$, each horizontal face of keyframe $i$ pairs with the closest rotated horizontal face of keyframe $j$ for 6-DoF optical-flow pose refinement. Across agreeing face pairs, confident flow correspondences are back-projected with $D_i$ and $D_j$ into 3D point pairs $\mathcal{M}_{ij}=\{(\mathbf{p},\mathbf{q})\}$ for weighted Umeyama alignment:
\begin{equation}
\tilde{\mathbf{S}}_{ij}=\arg\min_{\mathbf{S}\in\mathrm{Sim}(3)}\sum_{(\mathbf{p},\mathbf{q})\in\mathcal{M}_{ij}}\omega_{pq}\,\big\|\mathbf{p}-\mathbf{S}\,\mathbf{q}\big\|^{2},
\label{eq:umeyama}
\end{equation}
with flow weights $\omega_{pq}$, recovering relative scale $\tilde{s}_{ij}$, rotation $\tilde{\mathbf{R}}_{ij}$, and point-cloud translation $\mathbf{t}^{\mathrm{um}}_{ij}$ at any orientation.

\noindent\textbf{Global $\mathrm{Sim}(3)$ Consensus.}\label{sec:cycle} Matching over all 3D rotations also produces false matches in symmetric scenes, where a rigid symmetry $\mathbf{G}\neq\mathbf{I}$ such as a $180^\circ$ rotation maps a room or corridor onto itself. Transforming each candidate $(i,j)$ through local poses gives an inter-agent similarity $\hat{\mathbf{S}}_{ij}=\mathbf{T}_i\tilde{\mathbf{S}}_{ij}\mathbf{T}_j^{-1}$ from agent $b$ to agent $a$, grouping true closures across distinct rooms around one transform $\bar{\mathbf{S}}_{ab}$ while isolated room symmetries scatter across offsets $\bar{\mathbf{S}}_{ab}\mathbf{G}^{-1}$. When an entire shared region obeys one symmetry $\mathbf{G}$, however, false closures form a second cluster $\bar{\mathbf{S}}'_{ab}=\bar{\mathbf{S}}_{ab}\mathbf{G}^{-1}$ that survives pairwise robust kernels. Over any closed three-agent loop $(a,b,c)$, true similarities compose to $\bar{\mathbf{S}}_{ab}\bar{\mathbf{S}}_{bc}\bar{\mathbf{S}}_{ca}=\mathbf{I}$, whereas substituting $\bar{\mathbf{S}}'_{ab}$ gives
\begin{equation}
\bar{\mathbf{S}}'_{ab}\,\bar{\mathbf{S}}_{bc}\,\bar{\mathbf{S}}_{ca}=\bar{\mathbf{S}}_{ab}\,\mathbf{G}^{-1}\,\bar{\mathbf{S}}_{ab}^{-1}\neq\mathbf{I},
\label{eq:cycle}
\end{equation}
whose unknown scales cancel and whose rotation angle equals that of $\mathbf{G}$, far exceeding slow drift.

We therefore formulate outlier rejection as global consensus maximization over $\mathrm{Sim}(3)$ clusters. Grouping $\{\hat{\mathbf{S}}_{ij}\}$ of each pair $(a,b)$ by translation and rotation into clusters $\mathcal{H}_{ab}=\{(\bar{\mathbf{S}}^{h}_{ab},n^{h}_{ab})\}$ with robust mean $\bar{\mathbf{S}}^{h}_{ab}$ and support $n^{h}_{ab}$, we select at most one cluster $h_{ab}\in\mathcal{H}_{ab}\cup\{\varnothing\}$ ($n^{\varnothing}_{ab}=0$):
\begin{equation}
\{h^{*}_{ab}\}=\arg\max_{\{h_{ab}\}}\ \sum_{(a,b)}n^{h_{ab}}_{ab}\ \ \text{s.t.}\ \  \bar{\mathbf{S}}^{h_{ab}}_{ab}\,\bar{\mathbf{S}}^{h_{bc}}_{bc}\,\bar{\mathbf{S}}^{h_{ca}}_{ca}\approx\mathbf{I},
\label{eq:select}
\end{equation}
meaning loop residuals satisfy $\|\mathbf{t}_{abc}\|\le\tau_t$ and $\|\mathrm{Log}(\mathbf{R}_{abc})\|\le\tau_R$ on every triple $(a,b,c)$. Exhaustive search over top pair modes solves Eq.~\eqref{eq:select} exactly, rejecting symmetric clusters on three-agent loops while keeping the largest multi-room cluster on two-agent or tree pairs \cite{mangelson2018pcm,forsgren2022groupk}.

\subsection{Directional Global Optimization}
\label{sec:pgo}\label{sec:opt}
The loop closures $\mathcal{L}$ retained in $\{h^{*}_{ab}\}$ connect agents at multiple places along their trajectories, where unobservable monocular scale accumulates into drift \cite{strasdat2010scale}. Assigning one similarity per agent cannot align a trajectory whose scale changes across rooms, so each keyframe $i\in\mathcal{V}$ receives its own similarity $\mathbf{S}_i=(s_i,\mathbf{R}_i,\mathbf{t}_i)\in\mathrm{Sim}(3)$ mapping its camera frame to the global frame. Given a relative measurement $(\tilde{s}_{ij},\tilde{\mathbf{R}}_{ij},\tilde{\mathbf{t}}_{ij})$ on edge $(i,j)$, we compare $\mathbf{S}_i^{-1}\mathbf{S}_j$ against that measurement via the residual:
\begin{equation}
\mathbf{r}_{ij}=\begin{bmatrix}\mathrm{Log}\big(\tilde{\mathbf{R}}_{ij}^{\top}\mathbf{R}_i^{\top}\mathbf{R}_j\big)\\[2pt]s_i^{-1}\mathbf{R}_i^{\top}(\mathbf{t}_j-\mathbf{t}_i)-\tilde{\mathbf{t}}_{ij}\\[2pt]\log s_j-\log s_i-\log\tilde{s}_{ij}\end{bmatrix},
\label{eq:res}
\end{equation}
where evaluating translation error in the local frame and scale $s_i$ of keyframe $i$ keeps agents initialized at different monocular scales balanced. Summing over consecutive intra-agent odometry edges $\mathcal{O}$ and verified inter-agent loop closures $\mathcal{L}$ defines the pose-graph objective:
\begin{equation}
E=\sum_{(i,j)\in\mathcal{O}\cup\mathcal{L}}w_{ij}\,\big\|\mathbf{r}_{ij}\big\|^{2}_{\boldsymbol{\Omega}_{ij}},
\label{eq:pgo}
\end{equation}
with information matrix $\boldsymbol{\Omega}_{ij}$ and edge weight $w_{ij}$, where $w_{ij}=1$ on $\mathcal{O}$.

\noindent\textbf{Per-Keyframe Scale.} Inter-agent closures in $\mathcal{L}$ observe relative scale only where two trajectories meet, leaving keyframes between meetings underconstrained. To constrain them, on each odometry edge $(i,j)\in\mathcal{O}$ between consecutive keyframes of an agent, we take $(\tilde{\mathbf{R}}_{ij},\tilde{\mathbf{t}}_{ij})$ from the tracked motion $\mathbf{T}_i^{-1}\mathbf{T}_j$ and set $\tilde{s}_{ij}=1$. The third row of Eq.~\eqref{eq:res} then reduces to the log-scale increment $\log s_j-\log s_i$, weighted in $\boldsymbol{\Omega}_{ij}$ by the inverse variance of per-step scale drift. Minimizing these squared increments models $\log s_i$ as a random walk along each trajectory, so the scale term alone interpolates log-scale linearly between meetings and propagates scale to trajectory ends.

\noindent\textbf{Loop Constraint.}\label{sec:meas} On each loop closure $(i,j)\in\mathcal{L}$, translation magnitude and translation direction come from two estimates with different error sources. Weighted Umeyama alignment in Eq.~\eqref{eq:umeyama} provides scale $\tilde{s}_{ij}$, rotation $\tilde{\mathbf{R}}_{ij}$, and baseline length $\ell_{ij}=\|\mathbf{t}^{\mathrm{um}}_{ij}\|$, yet depth errors shift $\mathbf{t}^{\mathrm{um}}_{ij}$ in proportion to scene depth and distort its direction when two agents pass at a small baseline. Conversely, dense optical flow between paired cube faces constrains relative pose through pixel reprojection, so scaling all depths by one factor scales the translation magnitude while leaving its unit direction $\mathbf{u}_{ij}\in\mathcal{S}^2$ unchanged. We therefore form the loop translation measurement $\tilde{\mathbf{t}}_{ij}=\ell_{ij}\mathbf{u}_{ij}$ and weight errors along and perpendicular to $\mathbf{u}_{ij}$ separately in the translation block of $\boldsymbol{\Omega}_{ij}$:
\begin{equation}
\boldsymbol{\Omega}^{t}_{ij}=\lambda^{\parallel}_{ij}\,\mathbf{u}_{ij}\mathbf{u}_{ij}^{\top}+\lambda^{\perp}_{ij}\big(\mathbf{I}-\mathbf{u}_{ij}\mathbf{u}_{ij}^{\top}\big),
\label{eq:omega}
\end{equation}
where $\lambda^{\parallel}_{ij}$ inverts squared baseline noise at $\nu_{ij}=\max(\ell_{ij},\nu_0)$ and Umeyama depth error $\varepsilon_{ij}\bar{d}_i$ at median depth $\bar{d}_i$, and $\lambda^{\perp}_{ij}$ scales inverse face-pair flow angle variance $\vartheta_{ij}^{2}$ by $\nu_{ij}^{-2}$. Since all $m_{ij}$ closures of a meeting share surface depth errors, scaling robust weights $w_{ij}$ \cite{agarwal2013dcs} by $1/m_{ij}$ balances long and brief meetings in Eq.~\eqref{eq:pgo}.

\begin{table*}[t]
\centering
\caption{Per-agent ATE (m) on the Holo360D multi-agent panoramic benchmark, grouped by front cube-face or full-panorama input. Mean averages the five development scenes, and All Test Scenes reports mean ATE and tracked rate over 45 test scenes, where $^\dagger$ averages tracked agents and -- marks untracked scenes. Dark and light orange mark 1st and 2nd among multi-agent methods.}
\label{tab:main}
\small
\renewcommand{\arraystretch}{1.02}
\setlength{\tabcolsep}{3.0pt}
\begin{tabular}{lccccc|c|ccccc|cc}
\toprule
 & \multicolumn{6}{c|}{Development Scenes} & \multicolumn{5}{c|}{Five Longest Test Scenes} & \multicolumn{2}{c}{All Test Scenes} \\
Method & 002 & 005 & 006 & 007 & 009 & Mean & 055 & 011 & 047 & 003 & 043 & ATE & Track \\
\midrule
\multicolumn{14}{l}{\textit{Single-Agent, Front Cube Face}} \\
DROID-SLAM \cite{teed2021droid} & 1.011 & 2.050 & 2.548 & 0.843 & 1.934 & 1.677 & 4.891 & 7.389 & 1.162 & 2.972 & 1.840 & 1.610$^\dagger$ & 99\% \\
DPVO \cite{teed2023dpvo} & 1.836 & 2.373 & 2.601 & 1.159 & 1.802 & 1.954 & 5.993 & 6.084 & 1.289 & 2.622 & 1.741 & 1.576 & 100\% \\
MASt3R-SLAM \cite{murai2025mast3rslam} & 0.771 & 0.236 & 0.232$^\dagger$ & 0.915 & 0.638 & 0.559$^\dagger$ & 3.671$^\dagger$ & 4.080 & 0.215 & 1.457 & 0.152 & 0.651$^\dagger$ & 88\% \\
TTT3R \cite{chen2026ttt3r} & 3.037 & 2.002 & 1.726 & 1.349 & 1.935 & 2.010 & 10.749 & 9.196 & 0.933 & 4.455 & 1.985 & 2.542$^\dagger$ & 99\% \\
VGGT-Long \cite{deng2025vggtlong} & 2.792 & 1.106 & 1.589 & 1.573 & 2.741 & 1.960 & 9.062 & 8.446 & 1.217 & 4.588 & 0.910 & 2.000$^\dagger$ & 97\% \\
\multicolumn{14}{l}{\textit{Single-Agent, Panorama}} \\
ODGS-SLAM \cite{spiss2026odgsslam} & 2.605 & 0.753 & 3.173$^\dagger$ & 1.942$^\dagger$ & 3.009 & 2.296$^\dagger$ & 3.609$^\dagger$ & -- & 2.216$^\dagger$ & 2.018 & 1.837$^\dagger$ & 1.613$^\dagger$ & 63\% \\
OpenVSLAM \cite{sumikura2019openvslam} & 2.376 & 0.231 & 0.180 & 0.349 & 0.556 & 0.738 & 4.901 & 2.993 & 0.529 & 4.010 & 0.066 & 0.543$^\dagger$ & 94\% \\
Ours (Single) & 0.152 & 0.145 & 0.330 & 0.026 & 0.041 & 0.139 & 0.521 & 0.242 & 0.089 & 0.288 & 0.099 & 0.150 & 100\% \\
\midrule
\multicolumn{14}{l}{\textit{Multi-Agent, Front Cube Face}} \\
CP-SLAM \cite{hu2023cpslam} & 5.540$^\dagger$ & 4.023$^\dagger$ & 2.635$^\dagger$ & 2.597$^\dagger$ & 4.357$^\dagger$ & 3.830$^\dagger$ & -- & 13.807$^\dagger$ & -- & 4.392$^\dagger$ & -- & 3.869$^\dagger$ & 19\% \\
MAGiC-SLAM \cite{yugay2025magic} & -- & -- & -- & -- & -- & -- & -- & -- & -- & -- & -- & -- & 0\% \\
MAC-Ego3D \cite{xu2025macego3d} & -- & 4.151 & -- & -- & 4.232 & 4.191$^\dagger$ & 14.031 & 13.200 & -- & 5.172 & 2.314 & 4.080$^\dagger$ & 53\% \\
MAGS-SLAM \cite{mags2025} & \cellcolor{tabsecond}3.443 & 1.999$^\dagger$ & 2.839 & 1.063$^\dagger$ & 1.998 & 2.268$^\dagger$ & -- & -- & -- & 3.316$^\dagger$ & 2.067$^\dagger$ & 1.353$^\dagger$ & 83\% \\
CoMo3R-SLAM \cite{como3r} & 2.284$^\dagger$ & -- & -- & -- & 1.116 & 1.700$^\dagger$ & 0.691$^\dagger$ & \cellcolor{tabsecond}0.949 & 0.692$^\dagger$ & \cellcolor{tabsecond}1.525 & \cellcolor{tabsecond}0.159 & 0.606$^\dagger$ & 56\% \\
\multicolumn{14}{l}{\textit{Multi-Agent, Panorama}} \\
OpenVSLAM-M \cite{sumikura2019openvslam} & 3.743 & \cellcolor{tabsecond}0.443 & \cellcolor{tabsecond}0.184 & \cellcolor{tabsecond}0.972 & \cellcolor{tabsecond}0.581 & \cellcolor{tabsecond}1.184 & \cellcolor{tabsecond}4.473 & 3.073 & 0.382$^\dagger$ & 2.990 & 0.173 & 0.715$^\dagger$ & \cellcolor{tabsecond}88\% \\
Ours (Multi) & \cellcolor{tabfirst}0.023 & \cellcolor{tabfirst}0.019 & \cellcolor{tabfirst}0.013 & \cellcolor{tabfirst}0.014 & \cellcolor{tabfirst}0.037 & \cellcolor{tabfirst}0.021 & \cellcolor{tabfirst}0.155 & \cellcolor{tabfirst}0.161 & \cellcolor{tabfirst}0.053 & \cellcolor{tabfirst}0.300 & \cellcolor{tabfirst}0.034 & \cellcolor{tabfirst}0.082 & \cellcolor{tabfirst}100\% \\
\bottomrule
\end{tabular}
\end{table*}
\subsection{Omnidirectional Gaussian Fusion}
\label{sec:gsmap}
\begin{figure}[t]
\centering
\includegraphics[width=1\linewidth]{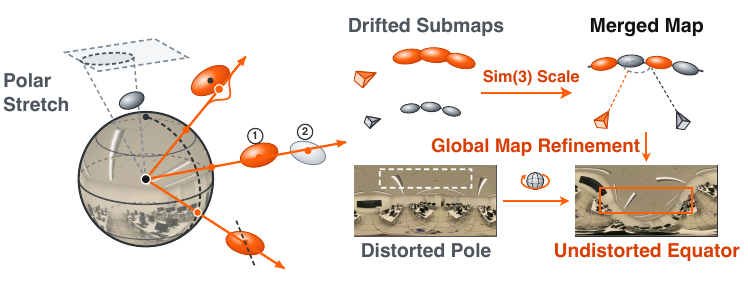}
\caption{Omnidirectional Gaussian fusion. Left: evaluating Gaussians at their maximum ray response and compositing front to back avoids polar stretch and dashed seam cuts. Right: per-keyframe $\mathrm{Sim}(3)$ scaling forms the merged map while random 3D rotations move poles onto the equator for 2D SSIM supervision.}
\label{fig:ray}
\end{figure}
With the keyframe similarities $\{\mathbf{S}_i\}_{i\in\mathcal{V}}$ solved by Eq.~\eqref{eq:pgo}, merging the local Gaussian maps into one panoramic map must account for both equirectangular distortion near the poles and varying scale $s_i$ along each trajectory. In Figure~\ref{fig:ray}, omnidirectional Gaussian fusion therefore renders Gaussians along 3D viewing rays, transforms each Gaussian with the similarity of the keyframe that created it, and jointly refines poses and Gaussians across all panoramas.

\noindent\textbf{Rendering Along Rays.} Standard pinhole rasterizers project each 3D Gaussian onto a tangent plane by local linearization \cite{kerbl20233dgs}, introducing distortion near the poles and discontinuities across the seam of an equirectangular image. Following ray-based Gaussian evaluation \cite{yu2024gof,moenne2024gaussianrt}, we represent each panoramic pixel directly by its unit viewing ray $\mathbf{d}\in\mathcal{S}^2$ and regularize each camera-frame Gaussian $(\boldsymbol{\mu},\boldsymbol{\Sigma},o)$ with a low-pass filter $\mathbf{A}=(\boldsymbol{\Sigma}+\kappa^{2}\|\boldsymbol{\mu}\|^{2}\mathbf{I})^{-1}$ scaled by pixel angular width $\kappa$. Along the ray $\tau\mathbf{d}$ for $\tau>0$, the Gaussian reaches its point of maximum response at $\tau^{*}=\mathbf{d}^{\top}\mathbf{A}\boldsymbol{\mu}/(\mathbf{d}^{\top}\mathbf{A}\mathbf{d})$, giving closed-form ray opacity:
\begin{equation}
\alpha(\mathbf{d})=o\,\exp\Big(-\tfrac{1}{2}\Big(\boldsymbol{\mu}^{\top}\mathbf{A}\boldsymbol{\mu}-\frac{(\mathbf{d}^{\top}\mathbf{A}\boldsymbol{\mu})^{2}}{\mathbf{d}^{\top}\mathbf{A}\mathbf{d}}\Big)\Big),
\label{eq:ray}
\end{equation}
for $\mathbf{d}^{\top}\mathbf{A}\boldsymbol{\mu}>0$ and $\alpha(\mathbf{d})=0$ behind the camera. At $\tau^{*}$, analytical gradients with respect to $\boldsymbol{\mu}$, $\boldsymbol{\Sigma}$, and pose depend only on the ray offset $\boldsymbol{\mu}-\tau^{*}\mathbf{d}$, and front-to-back alpha compositing ordered by radial distance $\|\boldsymbol{\mu}\|$ renders every latitude and across the seam without projection distortion in both local agent mapping and global fusion.

\noindent\textbf{Map Merging.} Since $s_i$ varies along each trajectory after Eq.~\eqref{eq:pgo}, rigidly transforming a local map misaligns revisited walls. Each Gaussian is therefore mapped by $\mathbf{T}_i^{-1}$ to the camera frame of the keyframe $i$ that created it, giving $(\boldsymbol{\mu},\boldsymbol{\Sigma})$, and then to the global frame by $\mathbf{S}_i$:
\begin{equation}
\boldsymbol{\mu}'=s_i\mathbf{R}_i\boldsymbol{\mu}+\mathbf{t}_i,\quad \boldsymbol{\Sigma}'=s_i^{2}\,\mathbf{R}_i\boldsymbol{\Sigma}\mathbf{R}_i^{\top},
\label{eq:deform}
\end{equation}
where scaling $\boldsymbol{\Sigma}$ by $s_i^{2}$ scales ellipsoid axes by $s_i$, preserving aspect ratios before opacity pruning and voxel downsampling merge overlapping Gaussians into $\mathcal{G}$.

\noindent\textbf{Global Map Refinement.}\label{sec:refine} Photometric supervision across overlapping panoramas further provides dense sub-pixel alignment beyond sparse keyframe geometry. Holding the converged scales $s_i$ fixed, we jointly optimize the merged Gaussians $\mathcal{G}$ and keyframe poses $\mathbf{W}_i=(\mathbf{R}_i,\mathbf{t}_i)^{-1}\in\mathrm{SE}(3)$ by minimizing the solid-angle-weighted loss:
\begin{equation}
\mathcal{L}_{\mathrm{ref}}=\sum_{i,\mathbf{d}}\omega_{\mathbf{d}}\Big((1-\beta)\big|\hat{I}_{i}-I_{i}\big|+\beta\big(1-\mathrm{SSIM}_{i}\big)\Big),
\label{eq:refine}
\end{equation}
where $\hat{I}_{i}$ renders $\mathcal{G}$ at $\mathbf{W}_i$ via Eq.~\eqref{eq:ray}, $\omega_{\mathbf{d}}$ weights pixel $\mathbf{d}$ by its solid angle to balance polar and equatorial gradients, $\mathrm{SSIM}_{i}$ measures local structural similarity, and $\beta$ balances the two terms. Each step samples any agent keyframe and applies a shared random 3D rotation to its pose and panorama in Eq.~\eqref{eq:refine} so 2D $\mathrm{SSIM}_{i}$ windows avoid polar stretch and seam cuts while refitting view-dependent color.
\section{Experiments}
\label{sec:exp}
\begin{figure*}[t]
\centering
\includegraphics[width=1\linewidth]{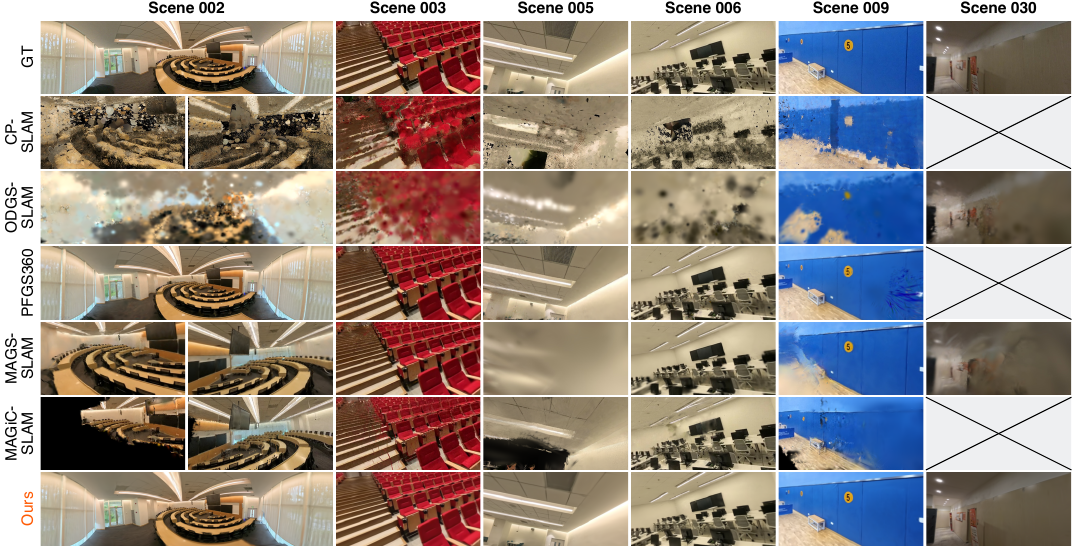}
\caption{Rendering comparison on four development and two test scenes. Each method renders a shared frame from its map at its estimated pose, showing the panorama on Scene 002 for panoramic methods, two cube faces for pinhole methods, and the front face elsewhere. Crosses mark failed runs, and black regions lie outside submaps.}
\label{fig:render}
\end{figure*}
\subsection{Setup}
\noindent\textbf{Benchmarks.} We build two multi-agent panoramic benchmarks from \textbf{Holo360D} \cite{holo360d} and \textbf{360Loc} \cite{huang2024loc360}. Following TopoMA \cite{zhang2026topoma}, 50 Holo360D indoor videos with ground-truth poses are split into four contiguous sub-sequences of length $\lfloor T/4\rfloor$ at 30\% boundary overlap (200 trajectories over 5 development and 45 test scenes), while 360Loc provides 4 to 5 independent day and night handheld recordings of 91\,m to 323\,m in each of four buildings (18 trajectories, no shared frames).

\noindent\textbf{Baselines.} We compare against single-agent and collaborative baselines with pinhole and panoramic inputs. On the $512\times512$ front cube face, DROID-SLAM \cite{teed2021droid}, DPVO \cite{teed2023dpvo}, MASt3R-SLAM \cite{murai2025mast3rslam}, TTT3R \cite{chen2026ttt3r}, and VGGT-Long \cite{deng2025vggtlong} evaluate independent tracking, whereas CP-SLAM \cite{hu2023cpslam}, MAGiC-SLAM \cite{yugay2025magic}, and MAC-Ego3D \cite{xu2025macego3d} (with Metric3D \cite{hu2024metric3dv2} depth), MAGS-SLAM \cite{mags2025}, and CoMo3R-SLAM \cite{como3r} evaluate collaborative loop closure and map merging. Full-panorama baselines include ODGS-SLAM \cite{spiss2026odgsslam} and its FPFH-RANSAC-ICP registration ODGS+RICP, pose-free PFGS360 \cite{zhuang2026pfgs360}, OpenVSLAM \cite{sumikura2019openvslam} in single-agent and sequential OpenVSLAM-M modes, and Ours (Single).

\noindent\textbf{Metrics.} We report per-agent ATE RMSE (m) under separate $\mathrm{Sim}(3)$ fits, joint ATE under one global $\mathrm{Sim}(3)$, and Track (completed runs). We mask operator pixels in tracking and mapping, and PSNR, SSIM, LPIPS, and WS-PSNR evaluate shared views at estimated poses (Table~\ref{tab:render}) and held-out non-keyframe panoramas every 5th frame (Table~\ref{tab:pano_gs}).

\noindent\textbf{Implementation Details.} Agents track six $90^\circ$ faces ($256\times256$) at 1.0 input FPS online in 7.8\,GiB GPU memory with frozen DROID-SLAM \cite{teed2021droid} and Metric3D \cite{hu2024metric3dv2} depth (12 initial keyframes), running 10 local $1024\times512$ mapping steps ($0.11$\,s/step) per new keyframe and sending $8.1$\,KB $128\times32$ ring descriptors for top-3 retrieval (13.8\,MB median traffic). On a $32\times64$ grid, tilts step by $7.5^\circ$ to $30^\circ$ plus $3^\circ$ refinement ($N_Q{=}108$), consensus keeps $\le 2$ clusters ($0.6$, $5^\circ$) under $\tau_t{=}1.0$, $\tau_R{=}5^\circ$ ($\le 3^{6}{=}729$ CPU states, $<1$\,ms), and pose graphs use $\sigma_R{=}1^\circ$, $\sigma_s{=}0.03$, DCS $\phi{=}2$, $\lambda^{\parallel}_{ij}=((\sigma_t\nu_{ij})^{2}+(\zeta\varepsilon_{ij}\bar{d}_i)^{2})^{-1}$, and $\lambda^{\perp}_{ij}=((\vartheta_{ij}^{2}+\sigma_R^{2})\nu_{ij}^{2})^{-1}$ ($\sigma_t{=}0.1$, $\zeta{=}0.281$, $\nu_0$ half median step), with offline alignment and 75k $2048\times1024$ steps (120k in Table~\ref{tab:pano_gs}, $\beta{=}0.1$, $0.32$\,s/step) on one A100 GPU.

\subsection{Comparison with Existing Methods}
\begin{table}[t]
\centering
\caption{Test trajectory comparison (ATE in m) after map refinement where each baseline tracks all agents. Single- and multi-agent baselines pair with Ours (Single) and Ours (Multi), Scenes counts valid scenes, and bottom rows cover all 45 scenes.}
\label{tab:test}
\small
\renewcommand{\arraystretch}{1.02}
\setlength{\tabcolsep}{1.4pt}
\begin{tabular}{lccccc}
\toprule
 & & \multicolumn{2}{c}{ATE} & \multicolumn{2}{c}{Joint} \\
Method & Scenes & Baseline & Ours & Baseline & Ours \\
\midrule
\multicolumn{6}{l}{\textit{Single-Agent}} \\
DROID-SLAM & 44 & 1.620 & \textbf{0.149} & -- & -- \\
DPVO & 45 & 1.576 & \textbf{0.150} & -- & -- \\
MASt3R-SLAM & 31 & 0.677 & \textbf{0.169} & -- & -- \\
TTT3R & 44 & 2.585 & \textbf{0.152} & -- & -- \\
VGGT-Long & 42 & 2.098 & \textbf{0.156} & -- & -- \\
ODGS-SLAM & 10 & 1.501 & \textbf{0.290} & -- & -- \\
OpenVSLAM & 38 & 0.611 & \textbf{0.157} & -- & -- \\
Ours (Single) & 45 & -- & \textbf{0.150} & -- & -- \\
\midrule
\multicolumn{6}{l}{\textit{Multi-Agent}} \\
MAC-Ego3D & 24 & 4.080 & \textbf{0.120} & 5.050 & \textbf{0.223} \\
MAGS-SLAM & 32 & 1.031 & \textbf{0.048} & 3.165 & \textbf{0.100} \\
CoMo3R-SLAM & 13 & 0.583 & \textbf{0.071} & 5.054 & \textbf{0.109} \\
OpenVSLAM-M & 35 & 0.824 & \textbf{0.083} & 1.882 & \textbf{0.134} \\
Ours (Multi) & 45 & -- & \textbf{0.082} & -- & \textbf{0.152} \\
\bottomrule
\end{tabular}
\end{table}
Table~\ref{tab:main} compares per-agent ATE on Holo360D. Benefiting from $360^\circ$ visibility, per-agent tracking and mapping redesigns tracking as a six-face rig with joint optical-flow bundle adjustment and initial-keyframe depth priors. Whereas pinhole trackers lose features during turns and RGB-D systems diverge under cross-room depth scale shifts, full-sphere constraints track 100\% of test agents at 0.150\,m ATE for Ours (Single), against 1.610\,m for DROID-SLAM, 1.576\,m for DPVO, and 19\% to 53\% tracked for CP-SLAM and MAC-Ego3D. On the server, spherical loop closure introduces $\mathrm{SO}(3)$ correlation and global $\mathrm{Sim}(3)$ consensus to match arbitrary headings and reject symmetric false clusters, while directional global optimization optimizes per-keyframe $\mathrm{Sim}(3)$ poses and decouples flow direction from scale, lowering 45-scene test ATE to 0.082\,m against 1.353\,m, 0.606\,m, and 0.715\,m for MAGS-SLAM, CoMo3R-SLAM, and \mbox{OpenVSLAM-M}.

\begin{table}[t]
\centering
\caption{Rendering quality of the front cube face on test scenes where each baseline outputs full or partial submaps, paired with our maps on shared frames over all pixels without masking.}
\label{tab:render}
\small
\renewcommand{\arraystretch}{1.02}
\setlength{\tabcolsep}{2.5pt}
\begin{tabular}{lccccc}
\toprule
 & & \multicolumn{2}{c}{PSNR} & \multicolumn{2}{c}{LPIPS} \\
Method & Scenes & Baseline & Ours & Baseline & Ours \\
\midrule
ODGS-SLAM & 43 & 17.67 & \textbf{27.92} & 0.620 & \textbf{0.231} \\
CP-SLAM & 19 & 20.85 & \textbf{27.60} & 0.602 & \textbf{0.241} \\
MAC-Ego3D & 12 & 16.32 & \textbf{29.36} & 0.658 & \textbf{0.214} \\
MAGiC-SLAM & 21 & 13.92 & \textbf{25.61} & 0.594 & \textbf{0.244} \\
MAGS-SLAM & 40 & 26.09 & \textbf{28.25} & 0.318 & \textbf{0.235} \\
\bottomrule
\end{tabular}
\end{table}
\begin{table}[t]
\centering
\caption{Omnidirectional 3DGS evaluation at $2048\times1024$ on the development set with operator masking: tracked agents, per-agent ATE (m), Scene 009 novel-view WS-PSNR and SSIM, and best single-agent or merged full-scene WS-PSNR over five scenes.}
\label{tab:pano_gs}
\footnotesize
\renewcommand{\arraystretch}{1.02}
\setlength{\tabcolsep}{1.1pt}
\begin{tabular}{lccccc}
\toprule
Method & Track & ATE & Novel WS & SSIM & Merged WS \\
\midrule
PFGS360 \cite{zhuang2026pfgs360} & 80\% & 0.212$^\dagger$ & 21.20 & 0.650 & -- \\
Ours (Single) & \cellcolor{tabfirst}100\% & 0.139 & \cellcolor{tabsecond}28.54 & \cellcolor{tabsecond}0.736 & \cellcolor{tabsecond}24.28 \\
w/o Global Map Refinement & \cellcolor{tabfirst}100\% & \cellcolor{tabsecond}0.055 & 18.23 & 0.598 & 17.81 \\
Ours (Multi) & \cellcolor{tabfirst}100\% & \cellcolor{tabfirst}0.021 & \cellcolor{tabfirst}28.95 & \cellcolor{tabfirst}0.745 & \cellcolor{tabfirst}27.80 \\
\bottomrule
\end{tabular}
\end{table}
\noindent\textbf{Pinhole Baselines.} Table~\ref{tab:test} pairs baselines with our system where they track all agents. While MAGS-SLAM and CoMo3R-SLAM miss opposite-heading views and align rigid submaps, spherical loop closure and directional global optimization match arbitrary orientations and optimize per-keyframe $\mathrm{Sim}(3)$ poses, reaching 0.048\,m and 0.071\,m ATE against 1.031\,m and 0.583\,m on 32 and 13 scenes.

\noindent\textbf{Panoramic Baselines.} Dense six-face optical flow in per-agent tracking and mapping preserves constraints in low-texture corridors and fast turns where sparse ORB matching in OpenVSLAM and photometric tracking in ODGS-SLAM lose lock, lowering ATE to 0.157\,m against 0.611\,m on OpenVSLAM's 38 scenes and 0.290\,m against 1.501\,m on ODGS-SLAM's 10 scenes. Unlike sequential relocalization onto a frozen map in OpenVSLAM-M, directional global optimization jointly optimizes all agents over verified closures to reach 0.083\,m ATE and 0.134\,m joint ATE against 0.824\,m and 1.882\,m on 35 scenes.

\begin{table}[t]
\centering
\caption{Multi-agent alignment wall-clock time (s) on Holo360D.}
\label{tab:time}
\footnotesize
\setlength{\tabcolsep}{3.5pt}
\begin{tabular}{lcccc}
\toprule
Method & Scene 002 & Scene 005 & Scene 007 & Total \\
\midrule
ODGS+RICP \cite{spiss2026odgsslam} & 3,869 & 5,454 & 5,809 & 15,132 \\
Ours (Multi) & \cellcolor{tabfirst}1,260 & \cellcolor{tabfirst}867 & \cellcolor{tabfirst}986 & \cellcolor{tabfirst}3,113 \\
\bottomrule
\end{tabular}
\end{table}
\noindent\textbf{Map Consistency.} Benefiting from full-sphere views, spherical loop closure combines ring-descriptor retrieval with 2D tilt search and FFT yaw correlation over $\mathrm{SO}(3)$. Across development meetings within 1\,m, $\mathrm{SO}(3)$ correlation recovers 3,950 closures at 95.6\% precision and connects 69.7\% of meetings against 1,152 closures and 28.9\% for front-face matching. Across 45 test scenes, it links 23,382 revisits outside shared boundaries among 33,815 closures (55.3\% above $45^\circ$ heading, reaching 0.036\,m ATE from revisits alone against 0.035\,m on 12 scenes), lowering test joint ATE in Table~\ref{tab:test} to 0.100\,m and 0.109\,m against 3.165\,m and 5.054\,m for MAGS-SLAM and CoMo3R-SLAM. Offline server alignment (GPU retrieval, $\mathrm{SO}(3)$ search, flow verification, and 22\,s to 75\,s CPU solve) takes 3,113\,s in Table~\ref{tab:time} ($4.9\times$ faster than 15,132\,s for ODGS+RICP) and removes drift in Figure~\ref{fig:teaser}.

\noindent\textbf{Rendering Quality.} Table~\ref{tab:render} and Figure~\ref{fig:render} evaluate rendered views at estimated poses following MAGS-SLAM. Whereas rigid submap merging in MAGS-SLAM leaves double walls at drifted revisits and tangent-plane rasterization in ODGS-SLAM stretches polar Gaussians, omnidirectional Gaussian fusion deforms Gaussians with per-keyframe $\mathrm{Sim}(3)$ poses and refines shared surfaces along viewing rays with solid-angle weights, reaching 28.25\,dB PSNR and 0.235 LPIPS against 26.09\,dB and 0.318 for MAGS-SLAM, and 27.92\,dB against 17.67\,dB for ODGS-SLAM. In Table~\ref{tab:pano_gs}, ray initialization from flow poses and four-route fusion track 100\% of runs in 4.3\,GPU-hours per agent against 80\% in 27.9\,GPU-hours for PFGS360 \cite{zhuang2026pfgs360}, raising Scene 009 novel-view WS-PSNR to 28.95\,dB against 21.20\,dB and full-scene WS-PSNR to 27.80\,dB against 24.28\,dB for Ours (Single).

\begin{table}[t]
\centering
\caption{Cross-session mapping ATE (m) over 18 day and night handheld sessions on 360Loc \cite{huang2024loc360} ($^\dagger$ partial, -- unmerged).}
\label{tab:360loc}
\footnotesize
\renewcommand{\arraystretch}{1.02}
\setlength{\tabcolsep}{1.8pt}
\begin{tabular}{lc|cccc|c}
\toprule
 & & \multicolumn{4}{c|}{Joint ATE (m)} & \\
Method & ATE & Atrium & Conc. & Hall & Piat. & Track \\
\midrule
Ours (Single) & 0.88 & -- & -- & -- & -- & 100\% \\
\midrule
MAC-Ego3D \cite{xu2025macego3d} & \cellcolor{tabsecond}23.59 & 20.69 & 28.30 & \cellcolor{tabsecond}29.43 & 34.07 & \cellcolor{tabfirst}100\% \\
MAGS-SLAM \cite{mags2025} & 2.04$^\dagger$ & \cellcolor{tabsecond}14.41 & 20.10 & -- & 34.62 & \cellcolor{tabsecond}83\% \\
ODGS+RICP \cite{spiss2026odgsslam} & 16.39$^\dagger$ & -- & -- & -- & \cellcolor{tabsecond}33.66 & 67\% \\
OpenVSLAM-M \cite{sumikura2019openvslam} & 0.38$^\dagger$ & -- & \cellcolor{tabfirst}0.14 & -- & -- & 72\% \\
Ours (Multi) & \cellcolor{tabfirst}0.30 & \cellcolor{tabfirst}0.22 & \cellcolor{tabsecond}0.35 & \cellcolor{tabfirst}0.44 & \cellcolor{tabfirst}0.39 & \cellcolor{tabfirst}100\% \\
\bottomrule
\end{tabular}
\end{table}
\noindent\textbf{Cross-Session Mapping on 360Loc.} Table~\ref{tab:360loc} evaluates cross-session mapping on 360Loc \cite{huang2024loc360} across four buildings under day and night lighting without shared frames. Although sparse ORB relocalization in OpenVSLAM-M reaches 0.14\,m on the lit shopfronts of Concourse, lighting changes cause it to fail on the other three buildings and track only 72\% of sessions. Even though our frozen DROID-SLAM weights were never trained on panoramic or day and night data, spherical loop closure verifies 912 to 3,014 closures per building via $\mathrm{SO}(3)$ correlation and global $\mathrm{Sim}(3)$ consensus, enabling directional global optimization and omnidirectional Gaussian fusion to align 100\% of sessions at 0.30\,m mean ATE against 0.88\,m for Ours (Single) and 0.22\,m to 0.44\,m joint ATE across all buildings against 14.41\,m to 34.62\,m for pinhole baselines.

\subsection{Ablation Study}
\begin{table}[t]
\centering
\caption{Ablation on shared six-face inputs (ATE / joint ATE in m, $^\dagger$ 3 scenes). Development and final test rows include refinement, while other test rows follow the pose graph.}
\label{tab:ablation}
\small
\renewcommand{\arraystretch}{1.02}
\setlength{\tabcolsep}{2.2pt}
\begin{tabular}{lcccc}
\toprule
 & \multicolumn{2}{c}{Development} & \multicolumn{2}{c}{Test} \\
Variant & ATE & Joint & ATE & Joint \\
\midrule
Ours (Single) & 0.139 & -- & 0.150 & -- \\
\midrule
w/ MAGS-SLAM Association & 0.172 & 1.329 & -- & -- \\
Pinhole Front-Face Only & 0.075 & 0.221 & 0.225 & 0.561 \\
w/o Cycle Consistency & 0.022 & 0.029 & 0.098 & 0.355 \\
w/o Per-Keyframe Scale & 0.092 & 0.162 & 0.144 & 0.245 \\
w/ Depth Translation Direction & 0.031 & 0.044 & 0.104 & 0.173 \\
w/ Projective Cube-Face 3DGS & 0.068$^\dagger$ & 0.077$^\dagger$ & -- & -- \\
w/o Global Map Refinement & 0.055 & 0.069 & 0.095 & 0.166 \\
\midrule
Ours (Multi) & \textbf{0.021} & \textbf{0.028} & \textbf{0.082} & \textbf{0.152} \\
\bottomrule
\end{tabular}
\end{table}
Table~\ref{tab:ablation} isolates each design on shared six-face trajectories and depth priors. Replacing our backend with MAGS-SLAM association degrades development ATE and joint ATE to 0.172\,m and 1.329\,m against 0.021\,m and 0.028\,m, isolating backend gains beyond $360^\circ$ views. In spherical loop closure, 2D tilt search and 1D FFT yaw correlation connect 62.2\% of meetings beyond $150^\circ$ (0.028\,m and 0.166\,m joint ATE against 0.083\,m for yaw-only search and 0.561\,m on front faces), while cycle-consistent global $\mathrm{Sim}(3)$ consensus rejects all 180 symmetric outliers above $10^\circ$ (108 $180^\circ$ flips) and keeps 99.3\% of true closures within 1.5\,m over 33,815 closures, reaching 0.166\,m against 0.355\,m and 0.314\,m and 0.264\,m against 7.545\,m and 0.673\,m on symmetric Scenes 001 and 021. In directional global optimization, random-walk log-scale regularization corrects 11\% median scale drift to 0.095\,m test ATE against 0.144\,m (median ATE stays 0.047\,m across $\sigma_s\in[0.01,0.03]$ against 0.056\,m at 0.1 and varies $\le 0.002$\,m over cluster radii $[0.3,1.0]$ and $\phi\in[1,5]$), and separating flow directions from depth magnitudes in Eq.~\eqref{eq:omega} reaches 0.021\,m development and 0.095\,m test ATE against 0.031\,m and 0.104\,m. In omnidirectional Gaussian fusion, ray evaluation avoids cube-seam gradients inflating six-face rasterization to 6.0M Gaussians against 0.50M (0.021\,m ATE and 27.18\,dB WS-PSNR against 0.068\,m and 22.80\,dB), and joint $\mathrm{Sim}(3)$ refinement reaches 0.021\,m ATE and 27.80\,dB WS-PSNR against 0.055\,m and 17.81\,dB, refining test ATE to 0.082\,m and 0.152\,m against 0.095\,m and 0.166\,m.
\section{Conclusion}
\label{sec:conclusion}
We proposed CoSphere-SLAM, the first multi-agent Gaussian splatting SLAM framework for panoramic cameras. Operating on $\mathcal{S}^2$, spherical loop closure matches keyframes at arbitrary orientations and rejects symmetric false closures, directional global optimization corrects scale drift along each trajectory, and omnidirectional Gaussian fusion merges local maps along viewing rays. Across indoor and cross-session panoramic benchmarks, CoSphere-SLAM achieves state-of-the-art trajectory accuracy and rendering quality without prior poses or scales. Although our system aligns panoramic agents across arbitrary headings and lighting changes, purely visual mapping still assumes static scenes and shared views. Future work will model dynamic objects on the sphere and integrate inertial and ranging sensors for metric scale and onboard deployment.

\bibliographystyle{IEEEtran}
\bibliography{main}

\begin{thebibliography}{10}
\providecommand{\url}[1]{#1}
\csname url@samestyle\endcsname
\providecommand{\newblock}{\relax}
\providecommand{\bibinfo}[2]{#2}
\providecommand{\BIBentrySTDinterwordspacing}{\spaceskip=0pt\relax}
\providecommand{\BIBentryALTinterwordstretchfactor}{4}
\providecommand{\BIBentryALTinterwordspacing}{\spaceskip=\fontdimen2\font plus
\BIBentryALTinterwordstretchfactor\fontdimen3\font minus \fontdimen4\font\relax}
\providecommand{\BIBforeignlanguage}[2]{{%
\expandafter\ifx\csname l@#1\endcsname\relax
\typeout{** WARNING: IEEEtran.bst: No hyphenation pattern has been}%
\typeout{** loaded for the language `#1'. Using the pattern for}%
\typeout{** the default language instead.}%
\else
\language=\csname l@#1\endcsname
\fi
#2}}
\providecommand{\BIBdecl}{\relax}
\BIBdecl

\bibitem{kerbl20233dgs}
B.~Kerbl, G.~Kopanas, T.~Leimk{\"u}hler, and G.~Drettakis, ``{3D} {G}aussian splatting for real-time radiance field rendering,'' \emph{ACM Transactions on Graphics}, vol.~42, no.~4, pp. 1--14, 2023.

\bibitem{keetha2024splatam}
N.~Keetha, J.~Karhade, K.~M. Jatavallabhula, G.~Yang, S.~Scherer, D.~Ramanan, and J.~Luiten, ``{SplaTAM}: Splat, track \& map {3D} {G}aussians for dense {RGB-D} {SLAM},'' in \emph{Proceedings of the IEEE/CVF Conference on Computer Vision and Pattern Recognition (CVPR)}, 2024, pp. 21\,357--21\,366.

\bibitem{matsuki2024monogs}
H.~Matsuki, R.~Murai, P.~H.~J. Kelly, and A.~J. Davison, ``{G}aussian splatting {SLAM},'' in \emph{Proceedings of the IEEE/CVF Conference on Computer Vision and Pattern Recognition (CVPR)}, 2024, pp. 18\,039--18\,048.

\bibitem{huang2024photoslam}
H.~Huang, L.~Li, H.~Cheng, and S.-K. Yeung, ``{Photo-SLAM}: Real-time simultaneous localization and photorealistic mapping for monocular, stereo, and {RGB-D} cameras,'' in \emph{Proceedings of the IEEE/CVF Conference on Computer Vision and Pattern Recognition (CVPR)}, 2024, pp. 21\,584--21\,593.

\bibitem{lee2024odgs}
S.~Lee, J.~Chung, J.~Huh, and K.~M. Lee, ``{ODGS}: {3D} scene reconstruction from omnidirectional images with {3D} {G}aussian splattings,'' in \emph{Advances in Neural Information Processing Systems (NeurIPS)}, 2024, pp. 57\,050--57\,075.

\bibitem{li2025omnigs}
L.~Li, H.~Huang, S.-K. Yeung, and H.~Cheng, ``{OmniGS}: Fast radiance field reconstruction using omnidirectional {G}aussian splatting,'' in \emph{Proceedings of the IEEE/CVF Winter Conference on Applications of Computer Vision (WACV)}, 2025, pp. 2260--2268.

\bibitem{mao2026panogsslam}
Y.~Mao, H.~Shi, Y.~Zhang, Z.~Yi, X.~Guo, and K.~Wang, ``{PanoGS-SLAM}: Panoramic {3D} {G}aussian splatting {SLAM},'' \emph{arXiv preprint arXiv:2609.17387}, 2026.

\bibitem{guo2026cubesplat}
X.~Guo, H.~Shi, Y.~Zhang, Z.~Yi, Y.~Mao, X.~Yin, and K.~Wang, ``{Cube-Splat}: High-fidelity 360$^\circ$ {G}aussian splatting {SLAM} via cubemap factorization and adjoint-consistent optimization,'' in \emph{Proceedings of the European Conference on Computer Vision (ECCV)}, 2026, pp. 213--231.

\bibitem{spiss2026odgsslam}
S.~Spiss, J.~Hieronimy, M.~Ritter, and M.~Harders, ``{ODGS-SLAM}: Omnidirectional {G}aussian splatting {SLAM},'' in \emph{Proceedings of the IEEE/CVF Conference on Computer Vision and Pattern Recognition (CVPR)}, 2026, pp. 26\,114--26\,123.

\bibitem{schmuck2019ccm}
P.~Schmuck and M.~Chli, ``{CCM-SLAM}: Robust and efficient centralized collaborative monocular simultaneous localization and mapping for robotic teams,'' \emph{Journal of Field Robotics}, vol.~36, no.~4, pp. 763--781, 2019.

\bibitem{lajoie2020door}
P.-Y. Lajoie, B.~Ramtoula, Y.~Chang, L.~Carlone, and G.~Beltrame, ``{DOOR-SLAM}: Distributed, online, and outlier resilient {SLAM} for robotic teams,'' \emph{IEEE Robotics and Automation Letters}, vol.~5, no.~2, pp. 1656--1663, 2020.

\bibitem{tian2022kimera}
Y.~Tian, Y.~Chang, F.~Herrera~Arias, C.~Nieto-Granda, J.~P. How, and L.~Carlone, ``{Kimera-Multi}: Robust, distributed, dense metric-semantic {SLAM} for multi-robot systems,'' \emph{IEEE Transactions on Robotics}, vol.~38, no.~4, pp. 2022--2038, 2022.

\bibitem{hu2023cpslam}
J.~Hu, M.~Mao, H.~Bao, G.~Zhang, and Z.~Cui, ``{CP-SLAM}: Collaborative neural point-based {SLAM} system,'' in \emph{Advances in Neural Information Processing Systems (NeurIPS)}, 2023, pp. 39\,429--39\,442.

\bibitem{deng2024mne}
T.~Deng, G.~Shen, C.~Xun, S.~Yuan, T.~Jin, H.~Shen, Y.~Wang, J.~Wang, H.~Wang, D.~Wang, and W.~Chen, ``{MNE-SLAM}: Multi-agent neural {SLAM} for mobile robots,'' in \emph{Proceedings of the IEEE/CVF Conference on Computer Vision and Pattern Recognition (CVPR)}, 2025, pp. 1485--1494.

\bibitem{yugay2025magic}
V.~Yugay, T.~Gevers, and M.~R. Oswald, ``{MAGiC-SLAM}: Multi-agent {G}aussian globally consistent {SLAM},'' in \emph{Proceedings of the IEEE/CVF Conference on Computer Vision and Pattern Recognition (CVPR)}, 2025, pp. 6741--6750.

\bibitem{xu2025macego3d}
X.~Xu, F.~Xue, S.~Zhao, Y.~Pan, S.~Scherer, and X.~Huang, ``{MAC-Ego3D}: Multi-agent {G}aussian consensus for real-time collaborative ego-motion and photorealistic {3D} reconstruction,'' in \emph{Proceedings of the IEEE/CVF Conference on Computer Vision and Pattern Recognition (CVPR)}, 2025, pp. 854--863.

\bibitem{mags2025}
Z.~Cao, Q.~Shao, S.~Zhai, J.~Zhang, A.~Nguyen, H.~Wang, and B.~Huang, ``{MAGS-SLAM}: Monocular multi-agent {G}aussian splatting {SLAM} for geometrically and photometrically consistent reconstruction,'' \emph{IEEE Transactions on Automation Science and Engineering}, 2026.

\bibitem{como3r}
Z.~Cao, Q.~Shao, S.~Zhai, F.~Tian, A.~Nguyen, H.~Wang, and B.~Huang, ``{CoMo3R-SLAM}: Collaborative monocular dense {SLAM} with learned {3D} reconstruction priors for outdoor multi-agent systems,'' \emph{arXiv preprint arXiv:2605.30488}, 2026.

\bibitem{schmuck2021covins}
P.~Schmuck, T.~Ziegler, M.~Karrer, J.~Perraudin, and M.~Chli, ``{COVINS}: Visual-inertial {SLAM} for centralized collaboration,'' in \emph{Proceedings of the IEEE International Symposium on Mixed and Augmented Reality Adjunct (ISMAR-Adjunct)}, 2021, pp. 171--176.

\bibitem{lajoie2024swarm}
P.-Y. Lajoie and G.~Beltrame, ``{Swarm-SLAM}: Sparse decentralized collaborative simultaneous localization and mapping framework for multi-robot systems,'' \emph{IEEE Robotics and Automation Letters}, vol.~9, no.~1, pp. 475--482, 2024.

\bibitem{mangelson2018pcm}
J.~G. Mangelson, D.~Dominic, R.~M. Eustice, and R.~Vasudevan, ``Pairwise consistent measurement set maximization for robust multi-robot map merging,'' in \emph{Proceedings of the IEEE International Conference on Robotics and Automation (ICRA)}, 2018, pp. 2916--2923.

\bibitem{yang2020gnc}
H.~Yang, P.~Antonante, V.~Tzoumas, and L.~Carlone, ``Graduated non-convexity for robust spatial perception: From non-minimal solvers to global outlier rejection,'' \emph{IEEE Robotics and Automation Letters}, vol.~5, no.~2, pp. 1127--1134, 2020.

\bibitem{agarwal2013dcs}
P.~Agarwal, G.~D. Tipaldi, L.~Spinello, C.~Stachniss, and W.~Burgard, ``Robust map optimization using dynamic covariance scaling,'' in \emph{Proceedings of the IEEE International Conference on Robotics and Automation (ICRA)}, 2013, pp. 62--69.

\bibitem{zach2010disambiguating}
C.~Zach, M.~Klopschitz, and M.~Pollefeys, ``Disambiguating visual relations using loop constraints,'' in \emph{Proceedings of the IEEE/CVF Conference on Computer Vision and Pattern Recognition (CVPR)}, 2010, pp. 1426--1433.

\bibitem{deng2025vggtlong}
K.~Deng, Z.~Ti, J.~Xu, J.~Yang, and J.~Xie, ``{VGGT-Long}: Chunk it, loop it, align it -- pushing {VGGT}'s limits on kilometer-scale long {RGB} sequences,'' in \emph{Proceedings of the IEEE International Conference on Robotics and Automation (ICRA)}, 2026, pp. 5446--5453.

\bibitem{chen2026ttt3r}
X.~Chen, Y.~Chen, Y.~Xiu, A.~Geiger, and A.~Chen, ``{TTT3R}: {3D} reconstruction as test-time training,'' in \emph{Proceedings of the International Conference on Learning Representations (ICLR)}, 2026.

\bibitem{sumikura2019openvslam}
S.~Sumikura, M.~Shibuya, and K.~Sakurada, ``{OpenVSLAM}: A versatile visual {SLAM} framework,'' in \emph{Proceedings of the ACM International Conference on Multimedia (MM)}, 2019, pp. 2292--2295.

\bibitem{huang2022360vo}
H.~Huang and S.-K. Yeung, ``{360VO}: Visual odometry using a single 360 camera,'' in \emph{Proceedings of the IEEE International Conference on Robotics and Automation (ICRA)}, 2022, pp. 5594--5600.

\bibitem{zhuang2026pfgs360}
C.~Zhuang, X.~Lu, Z.~Deng, Z.~Lu, Y.~Wang, J.~Diao, and J.~Xiao, ``Pose-free omnidirectional {G}aussian splatting for 360-degree videos with consistent depth priors,'' in \emph{Proceedings of the IEEE/CVF Conference on Computer Vision and Pattern Recognition (CVPR)}, 2026, pp. 4881--4890.

\bibitem{xu2022omniswarm}
H.~Xu, Y.~Zhang, B.~Zhou, L.~Wang, X.~Yao, G.~Meng, and S.~Shen, ``{Omni-swarm}: A decentralized omnidirectional visual-inertial-{UWB} state estimation system for aerial swarms,'' \emph{IEEE Transactions on Robotics}, vol.~38, no.~6, pp. 3374--3394, 2022.

\bibitem{xu2024d2slam}
H.~Xu, P.~Liu, X.~Chen, and S.~Shen, ``{D$^2$SLAM}: Decentralized and distributed collaborative visual-inertial {SLAM} system for aerial swarm,'' \emph{IEEE Transactions on Robotics}, vol.~40, pp. 3445--3464, 2024.

\bibitem{teed2021droid}
Z.~Teed and J.~Deng, ``{DROID-SLAM}: Deep visual {SLAM} for monocular, stereo, and {RGB-D} cameras,'' in \emph{Advances in Neural Information Processing Systems (NeurIPS)}, 2021, pp. 16\,558--16\,569.

\bibitem{forsgren2022groupk}
B.~Forsgren, R.~Vasudevan, M.~Kaess, T.~W. McLain, and J.~G. Mangelson, ``Group-$k$ consistent measurement set maximization for robust outlier detection,'' in \emph{Proceedings of the IEEE/RSJ International Conference on Intelligent Robots and Systems (IROS)}, 2022, pp. 4849--4856.

\bibitem{strasdat2010scale}
H.~Strasdat, J.~M.~M. Montiel, and A.~J. Davison, ``Scale drift-aware large scale monocular {SLAM},'' in \emph{Proceedings of Robotics: Science and Systems (RSS)}, 2010, pp. 73--80.

\bibitem{teed2023dpvo}
Z.~Teed, L.~Lipson, and J.~Deng, ``Deep patch visual odometry,'' in \emph{Advances in Neural Information Processing Systems (NeurIPS)}, 2023, pp. 39\,033--39\,051.

\bibitem{murai2025mast3rslam}
R.~Murai, E.~Dexheimer, and A.~J. Davison, ``{MASt3R-SLAM}: Real-time dense {SLAM} with {3D} reconstruction priors,'' in \emph{Proceedings of the IEEE/CVF Conference on Computer Vision and Pattern Recognition (CVPR)}, 2025, pp. 16\,695--16\,705.

\bibitem{yu2024gof}
Z.~Yu, T.~Sattler, and A.~Geiger, ``{G}aussian opacity fields: Efficient adaptive surface reconstruction in unbounded scenes,'' \emph{ACM Transactions on Graphics}, vol.~43, no.~6, pp. 1--13, 2024.

\bibitem{moenne2024gaussianrt}
N.~Mo{\"e}nne-Loccoz, A.~Mirzaei, O.~Perel, R.~de~Lutio, J.~Mart{\'\i}nez~Esturo, G.~State, S.~Fidler, N.~Sharp, and {\v{Z}}.~Gojcic, ``{3D} {G}aussian ray tracing: Fast tracing of particle scenes,'' \emph{ACM Transactions on Graphics}, vol.~43, no.~6, pp. 1--19, 2024.

\bibitem{holo360d}
J.~Ou, Z.~Cao, Y.~Ren, Z.~Li, J.~Zhu, T.~Hua, S.~Zhang, H.~Xiong, and W.~Zhao, ``{Holo360D}: A large-scale real-world dataset with continuous trajectories for advancing panoramic {3D} reconstruction and beyond,'' in \emph{Proceedings of the European Conference on Computer Vision (ECCV)}, 2026, pp. 682--699.

\bibitem{huang2024loc360}
H.~Huang, C.~Liu, Y.~Zhu, H.~Cheng, T.~Braud, and S.-K. Yeung, ``{360Loc}: A dataset and benchmark for omnidirectional visual localization with cross-device queries,'' in \emph{Proceedings of the IEEE/CVF Conference on Computer Vision and Pattern Recognition (CVPR)}, 2024, pp. 22\,314--22\,324.

\bibitem{zhang2026topoma}
X.~Zhang, S.~Shi, T.~Zhang, Z.~Guo, Z.~Huang, and Y.~Li, ``{TopoMA}: Topology-guided multi-agent dense {RGB} {3D} reconstruction via distributed inference,'' in \emph{Proceedings of the IEEE/CVF Conference on Computer Vision and Pattern Recognition (CVPR)}, 2026, pp. 21\,784--21\,793.

\bibitem{hu2024metric3dv2}
M.~Hu, W.~Yin, C.~Zhang, Z.~Cai, X.~Long, H.~Chen, K.~Wang, G.~Yu, C.~Shen, and S.~Shen, ``{Metric3D v2}: A versatile monocular geometric foundation model for zero-shot metric depth and surface normal estimation,'' \emph{IEEE Transactions on Pattern Analysis and Machine Intelligence}, vol.~46, no.~12, pp. 10\,579--10\,596, 2024.

\end{thebibliography}
\end{document}